\documentclass{article}
\usepackage{ijcai24}

\usepackage{times}
\usepackage{soul}
\usepackage{url}
\usepackage[hidelinks]{hyperref}
\usepackage[utf8]{inputenc}
\usepackage[small]{caption}
\usepackage{graphicx}
\usepackage{amsmath}
\usepackage{amsthm}
\usepackage{booktabs}
\usepackage{algorithm}
\usepackage{algorithmic}
\usepackage[switch]{lineno}

\usepackage{epsfig}
\usepackage{amssymb}
\usepackage{microtype}
\usepackage{pifont}
\usepackage{enumitem}
\usepackage{multirow}
\usepackage{makecell}
\newcommand{\cmark}{\ding{51}}
\newcommand{\xmark}{\ding{55}}
\usepackage{subfiles}

\newcommand{\equalcontrib}{\thanks{These authors contributed equally to this work. Author order is random.}}

\newcommand{\corrauthor}{\thanks{Corresponding author.}}

\title{Refined Temporal Pyramidal Compression-and-Amplification Transformer \\for 3D Human Pose Estimation}

\author{
    Hanbing Liu\textsuperscript{\rm 1,2,\equalcontrib},
    Wangmeng Xiang\textsuperscript{\rm 2,*}\corrauthor, 
    Jun-Yan He\textsuperscript{\rm 2,*}, 
    Zhi-Qi Cheng\textsuperscript{\rm 3,*}, \\
    Bin Luo\textsuperscript{\rm 2}, 
    Yifeng Geng\textsuperscript{\rm 2}, 
    Xuansong Xie\textsuperscript{\rm 2}
\affiliations
$^1$Tsinghua University\\
$^2$Institute for Intelligent Computing, Alibaba Group\\
$^3$Carnegie Mellon University\\
\emails
liuhb21@mails.tsinghua.edu.cn,
zhiqic@cs.cmu.edu, \\
\{wangmeng.xwm, leyuan.hjy, luwu.lb, cangyu.gyf\}@alibaba-inc.com,
xingtong.xxs@taobao.com
}

\begin{document}

\maketitle

\begin{abstract}
Accurate estimation of 3D human poses in video sequences demands both precision and a robust architectural framework. With the success of transformers, we introduce the \textbf{R}efined \textbf{T}emporal \textbf{P}yramidal \textbf{C}ompression-and-\textbf{A}mplification (RTPCA) transformer. Our approach addresses a critical issue in current transformer-based methods, which haven't fully harnessed the potential of intra\&inter-block attention mechanisms. Our Temporal Pyramidal Compression-and-Amplification (TPCA) module capitalizes on a temporal pyramid paradigm, significantly enhancing multi-scale key and value representations from intra-block. Realizing that many methods focus solely on single modules while overlooking their inter connections, we introduce the Cross-Layer Refinement (XLR) module. This module is designed to amplify inter-block communication, linking keys and values across adjacent blocks to create more coherent attention patterns. Integrating TPCA with XLR results in a powerful synergy, facilitating a rich semantic representation through the dynamic interaction of queries, keys, and values. The RTPCA's effectiveness is demonstrated by its remarkable performance on leading benchmarks like Human3.6M, HumanEva-I, and MPI-INF-3DHP, achieved with minimal computational overhead. The source code is available at \texttt{https://github.com/hbing-l/RTPCA}.
\end{abstract}

\begin{figure}[!t]
    \centering
    \centerline{\includegraphics[width=0.9\linewidth]{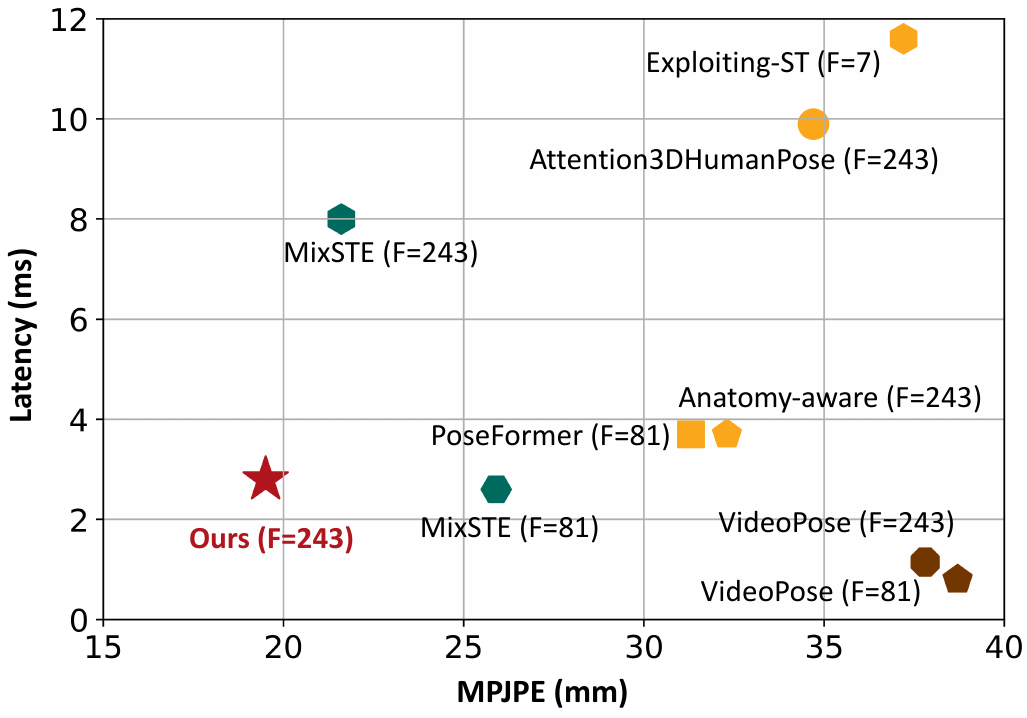}}
    \caption{\textbf{Comparison on the Human3.6M dataset to evaluate the performance of various methods in terms of MPJPE and latency.} Here the performance closer to the origin of the coordinates is considered more optimal. It indicates that our method (i.e., RTPCA) surpasses the others in terms of both accuracy and efficiency.}
    \vspace{-4mm}
    \label{fig:comparasion}
\end{figure}

\section{Introduction}

Deep learning architectures for 3D human pose estimation have witnessed exponential growth and success over the years~\cite{2021PoseAug,zheng(2021)-poseformer,hu2021conditional}. Notable strides have been made using body joint representations, human pose priors~\cite{hu2021conditional}, and establishing spatiotemporal relationships among frames~\cite{ZhangCVPR22MixSTE}. The wave of transformer-based attention mechanisms, specifically the Vision Transformer (ViT) \cite{Dosovitskiy_VIT_ICLR2021,chen2023hdformer}, has shown substantial results for 3D pose estimation.

Current transformer-based approaches often face challenges due to the limitations of attention mechanisms, which can result in a lack of diversity and depth in the information they yield. This issue significantly restricts the full potential of transformers in various applications. In response, certain approaches have endeavored to introduce more holistic information communication by employing cross-attention mechanisms \cite{tang20233d,li2022mhformer,hu2022acrnet}. However, these efforts have remained confined to intra-block contexts, lacking the exploration of multiscale information and inter-block interactions. Several ConvNet architectures such as HRNet \cite{sun2019deep}, EfficientDet \cite{tan2020efficientdet}, and others, have proven that the aggregation of neighboring layer features is incredibly effective in fusing spatial information and semantics. However, in transformer architectures, this feature aggregation scheme has not been fully explored.

To bridge this gap, we propose the Refined Temporal Pyramidal Compression-and-Amplification (RTPCA) strategy tailored for 3D pose estimation. Our framework is grounded in two primary components: The Temporal Pyramidal Compression-and-Amplification (TPCA) module and a cross-layer feature refinement (XLR) module. The TPCA module, in contrast to traditional U-Net~\cite{ronneberger2015u} and U-Net++~\cite{zhou2018unet++} architectures, dovetails feature refinement directly into attention operations. This module is strategically engineered to exploit temporal pyramidal structures, enabling the extraction of multi-scale information from within the intra-block attention mechanism. Similarly, our XLR module operates directly at the attention level, substantially boosting the model's efficiency in processing and integrating information. The XLR module allows later-stage transformer blocks to seamlessly access early-stage keys and values, boosting attention-driven learning and promoting interaction between inter-block modules. Furthermore, empirical results demonstrate that our approach is not only resistant to interference but also consistently delivers more stable outputs, indicating the robustness of our framework.

As illustrated in Figure~\ref{fig:comparasion}, we undertook a detailed comparative study on the Human3.6M dataset, assessing different methods based on their MPJPE and latency scores. MPJPE was calculated using ground truth 2D key points as input, while latency was measured in terms of inference speed at a batch size of one. We denoted the number of input frames as F. Notably, methods closer to the coordinate origin in our comparative chart were deemed more optimal. Our evaluation revealed that the RTPCA method outperformed competing approaches in terms of both accuracy and efficiency.

In summary, our contributions are as follows:
\begin{itemize}
\item \textbf{RTPCA Framework Development}: We unveil the RTPCA, a novel approach for 3D pose estimation. This approach addresses the inherent limitations of traditional transformer-based methods, efficiently facilitating the comprehensive exploitation of attention mechanisms and enhancing the model's robustness and stability.

\item \textbf{Compression-and-Amplification Design}: Our TPCA module, built upon a temporal pyramidal compression-and-amplification structure, is a novel solution for enhancing multi-scale key and value representations from intra-block.

\item \textbf{Cross-Layer Feature Refinement}: Our XLR module seamlessly interweaves feature refinement techniques with a cross-attention strategy. This novel integration method enables dynamic interactions across queries, keys, and values from inter-block layers.

\item \textbf{Superior Experimental Performance}: The effectiveness of RTPCA is validated against recognized benchmarks: Human3.6M, HumanEva-I, and MPI-INF-3DHP. Our model not only stands head and shoulders above its contemporaries in performance but also ensures computational efficiency.
\end{itemize}

\begin{figure*}[!t]
    \centering
    \centerline{\includegraphics[width=0.9\linewidth]{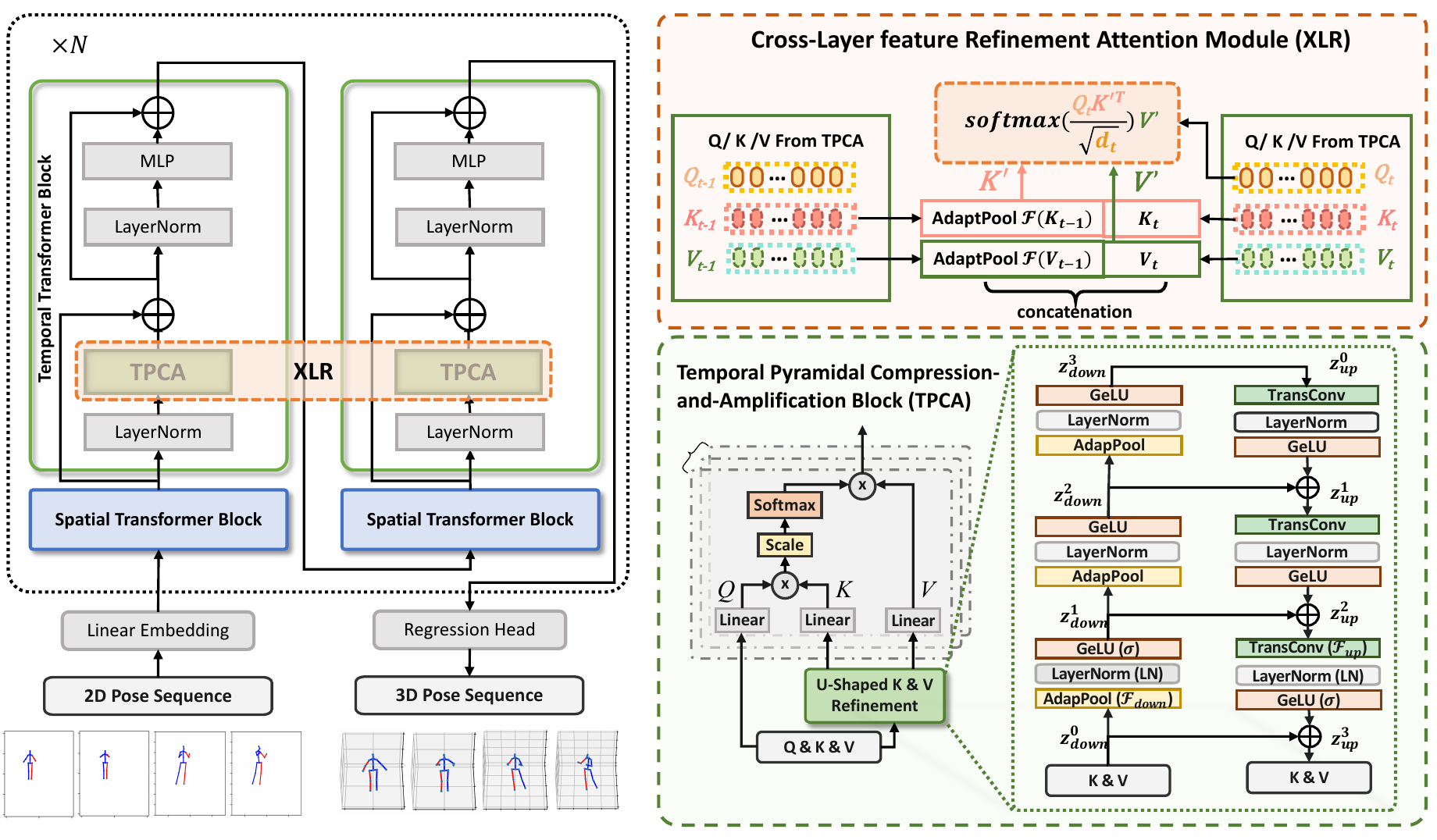}}
    \caption{\textbf{Framework of the proposed Refined Temporal Pyramidal Compression-and-Amplification (RTPCA).} The network is formed by stacking TPCA modules to extract multi-scale information in attention. The Cross-Layer Refinement (XLR) module is proposed to fuse inter-block information. The idea is to combine keys and values from both the front and the back for feature aggregation, thereby boosting the capability of the transformer. The input feature dimension equals $B\times J \times F \times C$, where $B$ denotes the batch size, $F$ is the number of frames, $J$ is the number of joints and $C$ represents the channel size.}
    \label{fig:pipeline}
\end{figure*}

\section{Related Work}
\noindent \textbf{3D Human Pose Estimation.} 3D Human Pose Estimation (HPE) has gained significant attention in computer vision, focusing on evaluating the 3D joint locations of a single human body from single or multiple views \cite{wang2021deep}. Approaches can be primarily categorized as one-stage and two-stage methods. One-stage approaches directly estimate the 3D pose from input images without intermediate 2D pose representation \cite{ma2021context}. In contrast, two-stage methods first detect 2D keypoints from the RGB data and then project these 2D keypoints to a 3D pose \cite{zhao2023poseformerv2,xu2021hourglass,tang20233d}. Considering the successful use of 2D keypoint detection in recent two-stage methods, our work adopts this method and focuses mainly on the 2D-to-3D lifting operation. 

\noindent \textbf{Graph Convolution Networks in 3D HPE.} Graph Convolution Networks (GCNs) \cite{kipf2016semi} have been extensively employed in HPE tasks due to their proficiency in encapsulating human skeleton graph information \cite{cai2019-exploiting}. For example, Wang \textit{et al.} \cite{wang2020motion} introduced a U-shaped GCN architecture and a novel motion loss to ensure smooth movement, while Hu \textit{et al.} \cite{hu2021conditional} proposed a U-shaped conditional directed graph convolutional network (U-CondDGCN). However, while these GCN-based approaches achieve impressive results in 3D HPE, they struggle to capture global contextual information across frames efficiently.

\noindent \textbf{Transformer in 3D HPE.} With the rise of Transformer architectures \cite{vaswani2017attention} and their potent self-attention mechanism, there has been a surge in its applications across a variety of vision tasks \cite{Dosovitskiy_VIT_ICLR2021,liu2023posynda}. For instance, PoseFormer \cite{zheng(2021)-poseformer} was introduced to model human joint relations within frames as well as temporal correlations across frames using a pure transformer structure. Similarly, other works proposed novel methods combining Transformer and GCN to effectively handle information flow interactions, or exploit connections within the network to improve the information exchange via cross-layer fusion \cite{chen2023hdformer}. However, these methods typically focus on enhancing individual modules and often overlook the interconnections among them. In contrast, our proposed TPCA method leverages the temporal pyramidal structure and cross-layer refinement mechanism to explore inter- and intra-block information within the attention module, thereby bolstering the transformer's capabilities.

\noindent \textbf{Inter\&Intra Attention Mechanism.}
Attention mechanisms are pivotal in transformer architectures, and an in-depth exploration of attention dynamics could significantly bolster transformer efficacy. Cross-attention stands as a commonly utilized mechanism for acquiring richer attention patterns. In \cite{tang20233d}, input features are bifurcated into temporal and spatial partitions, followed by concatenation after independent attention learning. \cite{li2022mhformer} fuses attention from multiple generated hypothesis, while \cite{hu2022acrnet} integrates features from different views. However, these methods confine fusion within a single block, disregarding inter-block connections, which can lead to significant discrepancies between the learned attention in a given block and its subsequent block. To address this, we propose an inter\&intra interaction mechanism, merging attention from adjacent stages to heighten model stability and robustness. As for intra-block exploration, numerous efforts have leveraged pyramidal networks to uncover multi-scale information. Inspired by \cite{wang2022uformpose} and \cite{yang2022u} which utilize the pyramidal architectures to the network structure and explore multi-scale feature attributes, we devise a pyramidal compression-and-amplification for attention Key \& Value, yielding a more comprehensive learning of attention information.

\section{Methodology}

\subsection{Framework}
The Refined Temporal Pyramidal Compression-and-Amplification (RTPCA) transformer model proposed in this study employs a transformer-based architecture to capture global patterns among input 2D human poses. The RTPCA framework, as shown in Figure \ref{fig:pipeline}, is composed of multiple Spatial-Temporal Encoders (STEs). Within this structure, each temporal transformer encoder is uniquely enhanced by integrating our specially designed TPCA module. The TPCA employs a pyramid compression-and-amplification model to derive multi-scale representations from keys and values. The adjacent TPCA modules are interconnected using the XLR module, which enables the combination of keys and values from consecutive temporal transformer encoders. This integration significantly enhances module interaction and promotes the efficient aggregation of information.

Formally, every frame of the input video is treated as a token in the transformer and represented as $X_{J, F} \in \mathbb{R}^{B \times J \times F \times 2}$. Here, $B$ denotes the batch size, $J$ refers to the number of joints, $F$ stands for the frames and 2 signifies the 2D coordinates of each joint. The input $X$ is initially projected into a high-dimensional feature space $Z_{J, F} \in \mathbb{R}^{B \times J \times F \times C}$, where $C$ signifies the channel size for each joint. Both temporal and spatial transformers are employed to alternately learn the spatiotemporal information in the input sequence, with the size of the input token reshaped as $Z_t \in \mathbb{R}^{(B \times J) \times F \times C}$ and $Z_s \in \mathbb{R}^{(B \times F) \times J \times C}$ respectively.

The transformer encoder takes $Z_t$ and $Z_s$ as input, and a regression head is followed to project the encoder's output to $Y_{J, F} \in \mathbb{R}^{B \times J \times F \times 3}$. The loss function of our proposed model is composed of three main components: Weighted Mean Per Joint Position Error (WMPJPE) $\mathcal{L}_w$, Temporal Coherence Loss (TCLoss) $\mathcal{L}_t$, and Mean Per Joint Velocity Error (MPJVE) $\mathcal{L}_m$. The WMPJPE $\mathcal{L}_w$ is formulated as $\mathcal{L}_w=\frac{1}{N} \sum_{i=1}^{N}\left(W \times \frac{1}{F} \sum_{j=1}^F\left\|p_{i, j}-g_{i, j}\right\|_2^2\right)$, where $N$ denotes $N$ joints of a human skeleton and $F$ represents the number of frames in a sequence. The terms $p_{i, j}$ and $g_{i, j}$ refer to the predicted and ground truth 3D poses of the $i$-th joint at the $j$-th frame, respectively. $W$ is a diagonal matrix consisting of weights for each joint, enabling us to give different importance to different joints. The overall loss function is given by $\mathcal{L}=\mathcal{L}_w+\lambda_t \mathcal{L}_t+\lambda_m \mathcal{L}_m$, where $\lambda_t$ and $\lambda_m$ are hyper-parameters for balancing the contributions of $\mathcal{L}_t$ and $\mathcal{L}_m$.
This design ensures the effective capture and representation of spatial-temporal human pose dynamics, thereby enhancing the performance of 3D human pose estimation.

\subsection{Preliminary}
The spatial-temporal dynamics of the human pose are typically modeled using two modules: the Spatial Transformer block (ST) and the Temporal Transformer block (TT). These modules are designed to learn the characteristics of human skeleton joints and the trajectories of different joints across frames, respectively.

\noindent \textbf{Spatial Transformer block.} The Spatial Transformer block (ST) captures the relationship of all joints within the human skeleton in each frame. The joint matrix of each frame is regarded as a spatial attention token $ Z_s \in \mathbb{R}^{(B \times F) \times J \times C} $. The tokens are then embedded with a spatial position matrix $ E_s \in \mathbb{R}^{(B \times F) \times J \times C} $, and fed into the transformer's basic components - multi-head self-attention (MSA) and multi-layer perceptron (MLP), as described in Vaswani et al. \cite{vaswani2017attention}. The dimension of the tokens remains unchanged after the feature extraction of ST. This process is formalized as follows:
\begin{equation}\label{equ:ST}
\begin{split}
\widetilde{\operatorname{ST}}(Z_s)&=Z_s + \operatorname{MSA}(Z_s), \\
\operatorname{ST}(Z_s)&=\widetilde{\operatorname{ST}}(Z_s)+\operatorname{MLP}(\widetilde{\operatorname{ST}}(Z_s)),
\end{split}
\end{equation}
In the first ST, the spatial position encoding $E_s$ is added to $Z_s$. $\operatorname{MSA}(\cdot)$ and $\operatorname{MLP}(\cdot)$ denote the operation of multi-head self-attention and multi-layer perceptron, respectively.

\noindent \textbf{Temporal Transformer block.} The Temporal Transformer block (TT), in conjunction with the novel XLR module, tracks the trajectory of each joint through the input frames. The joints are segmented into individual tokens $Z_t \in \mathbb{R}^{(B \times J) \times F \times C}$ in the temporal direction. Like ST, a time positional encoding $ E_t \in \mathbb{R}^{(B \times J) \times F \times C} $ is added to the input token, which is then fed into TT.
\begin{equation}\label{equ:TT}
\begin{split}
\widetilde{\operatorname{TT}}(Z_t) &= Z_t + \operatorname{XLR}(Z_{t-1}, Z_t), \\
\operatorname{TT}(Z_t)&=\widetilde{\operatorname{TT}}(Z_t)+\operatorname{MLP}(\widetilde{\operatorname{TT}}(Z_t)),
\end{split}
\end{equation}
In the equation above, $\operatorname{XLR}(\cdot)$ represents the operation of the cross-layer feature refinement module, accepting input tokens from both the previous block $Z_{t-1}$ and the current block $Z_t$, which will be explained in detail in Sec. \ref{sec:XA}.

\subsection{Temporal Pyramidal Compression-and- Amplification Attention Module}
\label{sec:UTA}

To achieve a more comprehensive extraction of information from the self-attention layer, we introduce a temporal pyramidal compression-and-amplification attention module (TPCA) to explore the multi-scale information embedded in keys $K \in \mathbb{R}^{n \times d}$ and values $V \in \mathbb{R}^{n \times d}$. The dimension of queries $Q \in \mathbb{R}^{n \times d}$ remains the same, while the keys and values are processed through multiple stages, thereby allowing the attention matrix to learn comprehensive pyramidal information while maintaining the same temporal resolution. As depicted in Figure~\ref{fig:pipeline}, this module iteratively compresses $K$ and $V$, reducing their sequence length by a factor of $r$ at each stage, with $r$ representing the sampling ratio. During the amplification phase, dimensions are gradually restored through the amplification layers, and residual links are introduced at corresponding phases to facilitate gradient flow.

\noindent \textbf{TPCA for attention Key \& Value.} Pyramidal networks have demonstrated their efficacy in various tasks owing to their ability to enhance robustness and capture high-level semantic information while preserving low-level details. To obtain more refined key and value representations, we leverage the temporal pyramidal compression-and-amplification network to enhance intra-block exploration. Given the input feature vector $z \in \mathbb{R}^{n \times d}$ with sequence length $n$ and channel dimension $d$, the TPCA module produces an output feature vector with the same dimensions. The operation of pyramidal temporal attention is represented as $\operatorname{TPCA}(\cdot)$. It can be formulated as follows:
\begin{equation}
\begin{split}
&{z}^{0}_{down} = z, \quad
{z}^{l+1}_{down}  = \sigma(\operatorname{LN}(\mathcal{F}_{down}({z}^l_{down}))), \\
&{z}^{0}_{up} = {z}^{m}_{down}, \quad
{z}^{l+1}_{up} = \sigma(\operatorname{LN}(\mathcal{F}_{up}({z}_{up}^{l}))) + {z}_{down}^{m-1-l}, \\
\end{split}
\label{eq:UTA}
\end{equation}
\noindent where $\sigma(\cdot)$ is the activation function, $\operatorname{LN}(\cdot)$ is the LayerNorm layer, and $\mathcal{F}_{down}$ and $\mathcal{F}_{up}$ denote the compression and amplification functions, respectively. ${z}^{l}_{down} \in \mathbb{R}^{\frac{n}{r^{l}} \times d}$ and ${z}^{l}_{up} \in \mathbb{R}^{\frac{n}{r^{m-l}} \times d}$. $l \in [0, 1, ..., m-1]$ is the index of the sampling stage. $r$ denotes the pooling ratio. The TPCA module first compresses the feature $z$ for $m$ stages, then amplifies the feature with a pyramidal connection. The final output $\operatorname{TPCA}(z) = {z}_{up}^{m} \in \mathbb{R}^{n \times d}$.


\begin{table*}[htb]
  \normalsize
  \centering
  \caption
  {
  \textbf{Quantitative results on Human3.6M.} The comparisons with SOTA methods on Human3.6M under protocol \#1 (top), protocol \#2 (middle) using detected 2D poses, and protocol \#1 (bottom) using ground truth 2D poses as input are conducted, where methods marked with $\dagger$ are video-based; CPN and GT denote the input 2D poses are estimated by CPN and ground truth respectively. The best and the second best results are highlighted in bold and underlined respectively.
  } 
  \resizebox{\textwidth}{!}{
  \begin{tabular}{@{}l|ccccccccccccccc|c@{}}
  \toprule
  \textbf{Protocol \#1 (CPN)} & Dir. & Disc & Eat & Greet & Phone & Photo & Pose & Purch. & Sit & SitD. & Smoke & Wait & WalkD. & Walk & WalkT. & Avg.\\
  \midrule[0.5pt]
  
  Pavlakos \textit{et al.} (CVPR2018)\cite{pavlakos2018ordinal} & 48.5 & 54.4 & 54.4 & 52.0 & 59.4 & 65.3 & 49.9 & 52.9 & 65.8 & 71.1 & 56.6 & 52.9 & 60.9 & 44.7 & 47.8 & 56.2 \\
  Cai \textit{et al.} (ICCV2019)\cite{cai2019-exploiting}$(\dagger)$ & 44.6 & 47.4 & 45.6 & 48.8 & 50.8 & 59.0 & 47.2 & 43.9 & 57.9 & 61.9 & 49.7 & 46.6 & 51.3 & 37.1 & 39.4 & 48.8 \\
    Pavllo \textit{et al.} (CVPR2019)\cite{pavllo2019-3d}$(\dagger)$ & 45.2 & 46.7 & 43.3 & 45.6 & 48.1 & 55.1 & 44.6 & 44.3 & 57.3 & 65.8 & 47.1 & 44.0 & 49.0 & 32.8 & 33.9 & 46.8 \\
    Xu \textit{et al.} (CVPR2020)\cite{xu2020-deep}$(\dagger)$ & {37.4} & 43.5 & 42.7 & 42.7 & 46.6 & 59.7 & {41.3} & 45.1 & {52.7} & {60.2} & 45.8 & 43.1 & 47.7 & 33.7 & 37.1 & 45.6 \\
    Liu \textit{et al.} (CVPR2020)\cite{liu2020-attention}$(\dagger)$ & 41.8 & 44.8 & 41.1 & 44.9 & 47.4 & 54.1 & 43.4 & 42.2 & 56.2 & 63.6 & 45.3 & 43.5 & 45.3 & 31.3 & 32.2 & 45.1 \\
    Wang \textit{et al.} (ECCV2020)\cite{wang2020motion}$(\dagger)$  & 41.3 & 43.9 & 44.0 & 42.2 & 48.0 & 57.1 & 42.2 & 43.2 & 57.3 & 61.3 & 47.0 & 43.5 & 47.0 & {32.6} & 31.8 & 45.6 \\
    Hu \textit{et al.} (ACMMM2021)\cite{hu2021conditional}$(\dagger)$ & 38.0 & 43.3 & {39.1} & \underline{39.4} & 45.8 & 53.6 & 41.4 & 41.4 & 55.5 & 61.9 & 44.6 & 41.9 & 44.5 & 31.6 & {29.4} & 43.4 \\
    Xu \textit{et al.} (CVPR2021)\cite{xu2021hourglass} & 45.2 & 49.9 & 47.5 & 50.9 & 54.9 & 66.1 & 48.5 & 46.3 & 59.7 & 71.5 & 51.4 & 48.6 & 53.9 & 39.9 & 44.1 & 51.9 \\
    Zheng \textit{et al.} (ICCV2021)\cite{zheng(2021)-poseformer}$(\dagger)$ & 41.5 & 44.8 & 39.8 & 42.5 & 46.5 & 51.6 & 42.1 & 42.0 & 53.3 & 60.7 & 45.5 & 43.3 & 46.1 & 31.8  & 32.2 & 44.3 \\
     Li \textit{et al.} (CVPR2022)\cite{li2022mhformer}$(\dagger)$ &{39.2} &{43.1} &{40.1} &{40.9} &{44.9} &{51.2} &{40.6} &{41.3} &{53.5} &{60.3} &{43.7} &{41.1} &{43.8} &{29.8} &{30.6} &{43.0} \\
     Zhang \textit{et al.} (CVPR2022)\cite{ZhangCVPR22MixSTE}$(\dagger)$ & \underline{37.6}    & \underline{40.9} & \underline{37.3} & {39.7} & \underline{42.3} & \underline{49.9}    & \underline{40.1}    & \underline{39.8} & \underline{51.7}          & \textbf{55.0} & \underline{42.1} & \underline{39.8} & \underline{41.0} & \underline{27.9} & \underline{27.9} & \underline{40.9} \\
    
  \midrule

  Ours$(\dagger)$ & \textbf{36.2} &\textbf{38.9} & \textbf{36.7} & \textbf{37.3} & \textbf{41.7} & \textbf{48.3} & \textbf{36.6} & \textbf{39.5} & \textbf{51.3} & \underline{57.2} & \textbf{41.4}  & \textbf{39.5} & \textbf{40.6} & \textbf{27.3} & \textbf{27.3} & \textbf{40.1} \\

  \toprule
  \textbf{Protocol \#2 (CPN)} & Dir. & Disc & Eat & Greet & Phone & Photo & Pose & Purch. & Sit & SitD. & Smoke & Wait & WalkD. & Walk & WalkT. & Avg.\\
  \midrule

  Cai \textit{et al.} (ICCV2019)\cite{cai2019-exploiting}$(\dagger)$ & 35.7 & 37.8 & 36.9 & 40.7 & 39.6 & 45.2 & 37.4 & 34.5 & 46.9 & 50.1 & 40.5 & 36.1 & 41.0 & 29.6 & 33.2 & 39.0 \\
    Pavllo \textit{et al.} (CVPR2019)\cite{pavllo2019-3d}$(\dagger)$ & 34.1 & 36.1 & 34.4 & 37.2 & 36.4 & 42.2 & 34.4 & 33.6 & 45.0 & 52.5 & 37.4 & 33.8 & 37.8 & 25.6 & 27.3 & 36.5 \\
    Xu \textit{et al.} (CVPR2020)\cite{xu2020-deep}$(\dagger)$ & 31.0 & 34.8 & 34.7 & 34.4 & 36.2 & 43.9 & 31.6 & 33.5 & 42.3 & 49.0 & 37.1 & 33.0 & 39.1 & 26.9 & 31.9 & 36.2 \\
    Liu \textit{et al.} (CVPR2020)\cite{liu2020-attention}$(\dagger)$ & 32.3 & 35.2 & 33.3 & 35.8 & 35.9 & 41.5 & 33.2 & 32.7 & 44.6 & 50.9 & 37.0 & 32.4 & 37.0 & {25.2} & 27.2 & 35.6 \\
    Wang \textit{et al.} (ECCV2020)\cite{wang2020motion}$(\dagger)$  & 32.9 & 35.2 & 35.6 & 34.4 & 36.4 & 42.7 & 31.2 & 32.5 & 45.6 & 50.2 & 37.3 & 32.8 & 36.3 & 26.0 & 23.9 & 35.5 \\
    Hu \textit{et al.} (ACMMM2021)\cite{hu2021conditional}$(\dagger)$ & \underline{29.8} & 34.4 & 31.9 & \underline{31.5} & 35.1 & 40.0 & \underline{30.3} & {30.8} & 42.6 & 49.0 & 35.9 & 31.8 & 35.0 & 25.7 & {23.6} & 33.8 \\
    Zhang \textit{et al.} (CVPR2022)\cite{ZhangCVPR22MixSTE}$(\dagger)$  & {30.8}    & \underline{33.1} & \textbf{30.3} & {31.8} & \underline{33.1} & \underline{39.1}    & {31.1}    & \textbf{30.5} & \underline{42.5}    & \textbf{44.5} & \underline{34.0} & \underline{30.8} & \underline{32.7} & \underline{22.1} & \underline{22.9}    & \underline{32.6} \\

  \midrule
  
  Ours$(\dagger)$ & \textbf{29.6} & \textbf{32.0} & \underline{30.7} & \textbf{30.5} & \textbf{32.3} & \textbf{37.2}  & \textbf{28.6} & \underline{30.6} & \textbf{41.3} & \underline{47.2} & \textbf{33.2} & \textbf{30.3} & \textbf{31.7} &  \textbf{21.1} & \textbf{22.2} & \textbf{32.0} \\
 \toprule
  \textbf{Protocol \#1 (GT)} & Dir. & Disc & Eat & Greet & Phone & Photo & Pose & Purch. & Sit & SitD. & Smoke & Wait & WalkD. & Walk & WalkT. & Avg.\\
  \midrule

  Pavllo \textit{et al.} (CVPR2019)\cite{pavllo2019-3d}$(\dagger)$ &35.2 &40.2 &32.7 &35.7 &38.2 &45.5 &40.6 &36.1 &48.8 &47.3 &37.8 &39.7 &38.7 &27.8 & 29.5 &37.8 \\   

  Zeng \textit{et al.} (ECCV2020)\cite{zeng2020srnet}$(\dagger)$ &34.8 & 32.1 & 28.5 & 30.7 & {31.4} &{36.9} &35.6 &{30.5} &{38.9} &40.5 &32.5 &{31.0} &29.9 & {22.5} & 24.5 &32.0 \\

  Zheng \textit{et al.} (ICCV2021)\cite{zheng(2021)-poseformer}$(\dagger)$ &{30.0} &{33.6} &{29.9} &31.0 &{30.2} &{33.3} &{34.8} &31.4 &{37.8} &{38.6} &{31.7} &{31.5} &{29.0} &23.3 &{23.1} &{31.3} \\
  Li \textit{et al.} (CVPR2022)\cite{li2022mhformer}$(\dagger)$ &{27.7} &{32.1} &{29.1} &{28.9} &{30.0} &{33.9} &{33.0} &{31.2} &{37.0} &{39.3} &{30.0} &{31.0} &{29.4} &{22.2} &{23.0} &{30.5} \\

  Zhang \textit{et al.} (CVPR2022)\cite{ZhangCVPR22MixSTE}$(\dagger)$ & \underline{21.6} & \underline{22.0} & \underline{20.4} & \underline{21.0} & \underline{20.8} & \underline{24.3} & \underline{24.7} & \underline{21.9} & \underline{26.9} & \textbf{24.9} & \underline{21.2} & \underline{21.5} & \underline{20.8} & \underline{14.7} & \underline{15.7} & \underline{21.6}\\

\midrule
  Ours$(\dagger)$ &\textbf{18.5} &\textbf{19.5} &\textbf{18.5} &\textbf{19.1} &\textbf{19.6} &\textbf{22.3} &\textbf{21.6} &\textbf{19.1} &\textbf{26.2} &\underline{25.3} &\textbf{19.1} &\textbf{19.3} &\textbf{18.5} &\textbf{12.5} &\textbf{12.8} &\textbf{19.5} \\

  \bottomrule
  
  \end{tabular}
  }
  \label{table:h36m}
\end{table*}

\subsection{Cross-Layer Refinement Module}
\label{sec:XA}


Unlike previous ConvNets \cite{sun2019deep,tan2020efficientdet} that employed simple addition and concatenation for feature aggregation, our proposed XLR module utilizes the interaction between queries, keys, and values across two TPCA modules of neighboring STEs. The structure of the XLR module is demonstrated in Fig.\ref{fig:pipeline}. This design effectively and naturally extends feature fusion to the transformer network by leveraging the attention scheme. More formally, the operation of the cross-layer attention, represented by $\operatorname{XLR}(\cdot)$, can be formulated as,
\begin{equation} \label{eq:XA}
\small
\begin{split}
\operatorname{XLR} &=\operatorname{Attn}(Z_{t-1},Z_{t}) \\
&=\operatorname{Attn}(Q_{t}, K', V'),  \\
&=\operatorname{Softmax}\left(\frac{Q_{t} K'^{\top} }{\sqrt{d}} \right) V', \\ 
\end{split}
\end{equation}
\begin{equation}
\small
\label{eq:kv}
\begin{split}
K' &= \operatorname{Concat}(K_{t}, \mathcal{F}(K_{t-1})),\\ 
V'&=\operatorname{Concat}(V_{t}, \mathcal{F}(V_{t-1})), \\
\end{split}
\end{equation}

\noindent where $Z_{t-1}$ and $Z_{t}$ are the latent features of the previous and current block respectively and $Z_{t}$ can be projected into query, key, and value through different weight matrices. $Q_{t}$, $K_{t}$, and $V_{t}$ are the query, key, and value from the second TT. $K_{t-1}$ and $V_{t-1}$ are the keys and values from the first TT. $K'$ and $V'$ are derived from the equation \ref{eq:kv}, where $\operatorname{Concat}$ denotes the concatenation operation, $\mathcal{F}$ is the adaptive pooling operated on $K_{t-1}$ and $V_{t-1}$ from the first TT. Both $K_{t}, V_{t}$ and $K_{t-1}, V_{t-1}$ are obtained after undergoing the TPCA module.

\section{Experiment}

\subsection{Datasets and Metric}
For our experimentation, we used three key 3D pose estimation benchmark datasets: Human3.6M \cite{Ionescu_POSE_TPAMI14}, HumanEva-I \cite{sigal2010humaneva}, and MPI-INF-3DHP \cite{mehta2017monocular}. Human3.6M is widely considered the industry standard for 3D pose evaluation, boasting 3.6 million video frames captured from four synchronized cameras at 50 Hz, each positioned differently. The dataset encompasses 11 subjects executing 15 different actions. HumanEva-I, while smaller in comparison with fewer subjects and actions, is still a robust dataset for 3D pose estimation. Following the practice of \cite{pavllo2019-3d} and \cite{lee2018propagating}, we utilized all actions (Walk, Jog, Box) from subjects $S1$, $S2$, and $S3$ for our training. The MPI-INF-3DHP contains both indoor and outdoor environmental datasets, involving more diverse motions than Human3.6M and HumanEva-I. 

To maintain comparability with previous work~\cite{hu2021conditional,cai2019-exploiting,ZhangCVPR22MixSTE}, we adopted the Mean Per Joint Position Error (MPJPE) for Human3.6M and HumanEva-I, MPJPE, Percentage of Correct Keypoints(PCK) with the threshold of 150mm, and Area Under Curve (AUC) for a range of PCK thresholds for MPI-INF-3DHP as our evaluation metric.

\subsection{Implementation Details}
We implemented the proposed RTPCA using the PyTorch framework\footnote{https://pytorch.org}, and our experiments were performed on two NVIDIA TITAN V100 GPUs. For performance analysis, we utilized 2D keypoints derived from a 2D pose detector \cite{Chen2018CPN}, as well as from 2D ground truth. The model was optimized using the Adam optimizer \cite{kingma2014adam}, and we established the batch size, dropout rate, and activation function at 1024, 0.1, and GELU, respectively. Following the example set by \cite{ZhangCVPR22MixSTE}, we designated the input sequence length to 243 for Human3.6M and 27 for HumanEva-I and MPI-INF-3DHP. The weights assigned in WMPJPE were differentiated based on joint groups, attributing 1.0 to the torso, 1.5 to the head, 2.5 to the middle limb, and 4.0 to the terminal limb. The $\lambda_t$ and $\lambda_m$ were referred to \cite{ZhangCVPR22MixSTE}. To prevent overlapping frames between sequences, we applied a stride data sampling strategy, where the interval matches the input length. 

\subsection{Comparison with the State-of-the-Arts}

\begin{table}
    \small
    \centering
    \caption{\textbf{Frame, Parameter(M), Latency(ms) and MPJPE(mm) of different methods.} The best result is highlighted in bold.}
    \setlength{\tabcolsep}{0.5mm}{
    \begin{tabular}{l|cccc}
\toprule
\textbf{Method} & Frame & Param. & Latency & MPJPE  \\
\midrule
\cite{pavllo2019-3d} (CVPR2019) & 81 & 12.75 & \textbf{0.8} & 38.7 \\
\cite{pavllo2019-3d} (CVPR2019) & 243 & 16.95 & 1.15 & 37.8 \\
\cite{cai2019-exploiting} (ICCV2019) & 7 & 5.04 & 11.6 & 37.2 \\
\cite{liu2020-attention} (CVPR2020) & 243 & 11.25 & 9.9 & 34.7 \\
\cite{zheng(2021)-poseformer} (ICCV2021) & 81 & 9.6 & 3.7 & 31.3 \\
 \cite{chen2021anatomy} (TCSVT2021) & 243 & 59.18 & 3.7 & 32.3 \\
 \cite{li2022-exploiting} (TMM2022) & 351 & 4.34 & 9.5 & 28.5 \\
 \cite{ZhangCVPR22MixSTE} (CVPR2022) & 243 & 33.78 & 8 & 21.6 \\
 \cite{huang2022dh} (ECCV2022) & 27 & - & - & 37.9 \\
 \cite{azizi20223d} (ECCV2022) & 128 & \textbf{0.16} & - & 36.2 \\
 \cite{zhao2023poseformerv2} (CVPR2023) & 27 & - & - & 35.5 \\
\cite{tang20233d} (CVPR2023) & 243 & 18.91 & -  & 21.3 \\
\midrule
Ours  & 243 & 33.91 & 2.8 & \textbf{19.5} \\ 
\bottomrule
\end{tabular}}
    \label{table:para}
\end{table}

\noindent \textbf{Human3.6M Results.} To evaluate the efficacy of our proposed methodology, we conducted a comparative analysis with the leading state-of-the-art (SOTA) approaches utilizing the Human3.6M dataset\cite{Ionescu_POSE_TPAMI14}. For accuracy, we have referred to the performance metrics as reported in the respective original publications of these competing methods. A comprehensive comparison of our approach, encompassing both GCN-based and Transformer-based methods, is presented in Table \ref{table:h36m}.

Our method sets a new benchmark in 3D pose estimation, achieving a MPJPE of 19.5mm, surpassing the current best, MixSTE \cite{ZhangCVPR22MixSTE}, by 2.7mm with a similar parameter count. Further evaluation, as shown in Table \ref{table:h36m}, highlights our method's efficacy with detected 2D poses using CPN \cite{Chen2018CPN} as the detector. Our approach achieves 40.1mm MPJPE under protocol \#1 and 32.0mm MPJPE under protocol \#2, underscoring its robustness and precision.

Table \ref{table:para} provides a comprehensive overview of recent and cutting-edge video-based methodologies, encompassing a detailed comparison of their parameters, FLOPs, and inference latency on Human3.6M under protocol 1 on Human3.6M. Considering various factors holistically, our method shows superior performance with marginal parameter overhead and relatively fast latency.

\noindent \textbf{HumanEva-I Results.} To assess the generalization ability of our proposed method, we compared its performance with SOTA methods using the HumanEva-I benchmark. As shown in Table \ref{table:humaneva}, the MPJPE results on HumanEva-I validate our method's effectiveness and transferability, mirroring the trend observed on the Human3.6M dataset. Once again, our proposed model achieves SOTA performance.

\begin{table}
    \small
    \centering
    \caption{\textbf{Quantitative results on HumanEva-I dataset under protocol \#1.} The best result is highlighted in bold.}
    \setlength{\tabcolsep}{0.4mm}{
    \begin{tabular}{@{}l@{}|ccc|ccc|c}
\toprule
\textbf{Protocol \#1} & \multicolumn{3}{c}{Walk}                      & \multicolumn{3}{c}{Jog}                       & \textbf{Avg.}           \\
\midrule
p-LSTMs \cite{lee2018propagating} & 18.6 & 19.9 & 30.5 & 25.7 & 16.8 & 17.7 & 30.3 \\
PoseFormer \cite{zheng(2021)-poseformer}   & 16.3          & 11.0            & 47.1          & 25.0            & 15.2          & 15.1          & 21.6          \\
MixSTE \cite{ZhangCVPR22MixSTE} & 20.3 & 22.4 & 34.8 & 27.3 & 32.1 & 34.3 & 28.5 \\
\cite{diaz2023convformer} & 12.5 & 10.1 & 25.4 & 13.3 & 12.9 & 22.6 & 20.6\\
\midrule
Ours &  18.8  &    17.8      &    21.2     &    18.5     &     21.2    &     21.3     &    \textbf{19.1}        \\

\bottomrule
    
\end{tabular}}
    \label{table:humaneva}
\end{table}

\begin{table}
    \small
    \centering
    \caption{\textbf{Quantitative comparisons with SOTA methods on MPI-INF-3DHP with three metrics.} Best in bold, second best best underlined.}
    \setlength{\tabcolsep}{0.5mm}{
    \begin{tabular}{l|@{}c@{}c@{}c@{}}
\toprule
\textbf{Method}        & PCK[↑] & AUC[↑] & MPJPE[↓] \\ 
\midrule
    VNect \cite{mehta2017-vnect}   & 79.4 & 41.6 & - \\
    TrajectoryPose3D \cite{lin2019-trajectory}  & 83.6 & 51.4 & 79.8 \\
    UGCN \cite{wang2020motion} & 86.9  & 62.1 & 68.1 \\
    U-CondDGCN \cite{hu2021conditional} & \underline{97.9}   & {69.5} & \underline{42.5} \\
    PoseAug \cite{2021PoseAug} & 88.6 & 57.3 & 73.0 \\
    MixSTE~\cite{ZhangCVPR22MixSTE}  &{94.4}   & 66.5 & 54.9 \\
    MLP-Graph~\cite{hassan2023spatio} & 86.8 & 54.5 & - \\
    HSTFormer~\cite{qian2023hstformer} & 96.6 & \underline{70.5} & 44.5 \\
    \midrule
    Ours  & \textbf{98.8}  & \textbf{74.2}   & \textbf{40.5} \\ 
\bottomrule
    
\end{tabular}}
    \label{table:3dhp}
\end{table}

\noindent \textbf{MPI-INF-3DHP Results.} Table \ref{table:3dhp} shows a quantitative comparison of different SOTA methods for 3D human pose estimation on MPI-INF-3DHP dataset under PCK, AUC, and MPJPE metrics.
The proposed method achieves the best performance across all metrics, outperforming other approaches.
Considering that the MPI-INF-3DHP dataset encompasses diverse indoor and outdoor scenarios, the superior performance of our method highlights its adaptability to complex environments.
The ability to achieve better results across all metrics suggests that our method can effectively handle noise and variations present in real-world settings.

\subsection{Ablation Study}
To thoroughly ascertain the efficacy of each component and design strategy employed in our proposed model, we conducted extensive ablation experiments using the Human3.6M dataset under protocol 1 with 2D ground truth human pose as input.
\begin{table}
    \small
    \centering
    \caption{\textbf{Ablation study for each component used in our method.} The evaluation is conducted on Human3.6M with MPJPE(mm), Parameters(M), and FLOPs(G).}
    \setlength{\tabcolsep}{1mm}{
    \begin{tabular}{l|ccc|ccc}
    \toprule
        \textbf{Method} & STE  & TPCA  & XLR & MPJPE[↓] & Params & FLOPs \\
    \midrule
    Baseline  &  \cmark & \xmark &  \xmark & 21.7 & 33.78 & 139.0\\
              & \cmark &  \cmark  & \xmark& 20.0& 33.84 & 139.4 \\
    \midrule
    Ours     & \cmark &  \cmark  & \cmark  & \textbf{19.5} & 33.91 & 139.4 \\
\bottomrule 
    \end{tabular}}
    \label{table:ablation_component}
\end{table}

\begin{table}
    \small
    \centering
    \caption{\textbf{Ablation study for the noisy input and accelation error loss.} Acceleration error, MPJPE using noisy ground truth 2D keypoints as input, and MPJPE using ground truth 2D keypoints as input are showcased in this table. The best result is highlighted in bold.}
    \setlength{\tabcolsep}{1.35mm}{
    \begin{tabular}{l | c c c}
    \toprule
     \textbf{Method} & Accel[↓]  & MPJPE(noisy input)[↓] &  MPJPE[↓]  \\ 
    \midrule
    STE   & 22.4  & 30.8 & 21.7\\
    STE+TPCA & 10.7 & 23.3 & 20.0\\
    STE+XLR & 8.5 & 25.7 & 20.9\\
    RTPCA & \textbf{7.3} & \textbf{20.8} & \textbf{19.5} \\
    \bottomrule
    \end{tabular}}
    \label{table:noise_stable}
\end{table}

\noindent \textbf{Impact of Model Components.} We systematically assessed the individual contributions of our model's components by incrementally integrating them as shown in Table \ref{table:ablation_component}. Our baseline model was the STE, which is essentially the SOTA method \cite{ZhangCVPR22MixSTE}. The addition of the TPCA module resulted in a significant reduction of MPJPE by 1.7mm, emphasizing the effectiveness of extracting multi-scale feature information. Furthermore, the implementation of the XLR module led to an improved pose estimation accuracy without a substantial increase in the parameters and FLOPs.

To assess noise reduction and stability enhancement, we introduced Gaussian noise to ground truth 2D human keypoints. The results in Table \ref{table:noise_stable} highlight our approach's robustness against perturbations. Notably, the TPCA module plays a significant role in enhancing the model's resilience. This is further augmented by the beneficial impact of the XLR module. To evaluate stability, we report the acceleration error (Accel), measuring the difference between the predicted and ground truth 3D acceleration for each keypoint in $mm/s^2$. Notably, the major enhancement in stability observed in our RTPCA method, as indicated by the lowest Accel value, is primarily due to the substantial contribution of the XLR module.

\begin{table}
    \small
    \centering
    \caption{\textbf{Ablation study for the structure of TPCA.} Different compression and amplification methods are analyzed in this table.}
    \setlength{\tabcolsep}{1.35mm}{
    \begin{tabular}{c c c c | c }
    \toprule
    \multicolumn{2}{c}{\textbf{\textit{Compression}}} & \multicolumn{2}{c}{\textbf{\textit{Amplification}}} & \multirow{2}{*}{MPJPE[↓]} \\
    \cmidrule(r){1-2}  \cmidrule(r){3-4} 
    Pool & Conv & Linear & Trans-Conv & \\
    \midrule
    \cmark & & & & 20.9 \\
     & \cmark & & & 22.1 \\
     \cmark & \cmark & & & 20.6 \\
     \cmark & & & \cmark & \textbf{20.0} \\
     \cmark & \cmark & \cmark & & 20.1 \\
     \cmark & \cmark & & \cmark & 20.9 \\

\bottomrule 
    \end{tabular}}
    \label{table:UTA}
\end{table}

\noindent \textbf{Impact of TPCA Module.} We explored two compression methods, adaptive pooling and convolution, and two amplification approaches, linear and transpose convolution, for comparative analysis. As illustrated in Table \ref{table:UTA}, the combination of compression and amplification (pyramidal structure) demonstrated superior performance over using compression alone, as evident from the improved results in the last three rows compared to the first three. Among the three different combination strategies tested, adaptive pooling paired with transpose convolution yielded the best performance, recording an MPJPE of 20.0mm. This combination was subsequently adopted in other experiments.

\begin{table}
    \small
    \centering
    \caption{\textbf{Ablation study for the cross-layer fusion strategies.} Feature-based and Attention-based methods are conducted in this experiment to explore the performance of different cross-layer fusion strategies.}
    \setlength{\tabcolsep}{1mm}{
    \begin{tabular}{c | c | c}
    \toprule
      \textbf{Method} & Structure  & MPJPE \\ 
    \midrule
     Baseline & TPCA & 20.0 \\
    \midrule
   \multirow{2}{*}{Feature-based} & TPCA+CNN & 21.7 \\
    & TPCA+MLP & 20.9 \\
    \midrule
    \multirow{3}{*}{Attention-based} &TPCA+Pyramidal XLR (Pyramidal $Q$)   &  21.2\\
    &TPCA+Pyramidal XLR (Constant $Q$)  & 19.9\\
    &TPCA+Residual XLR (Constant $Q$) & \textbf{19.5} \\
    \bottomrule
    \end{tabular}}
    \label{table:cross}
\end{table}

\noindent \textbf{Impact of Cross-Layer Connections.} The experimental results from various fusion strategies presented in Table \ref{table:cross} clearly demonstrate that attention-based methods for integrating information are more effective in harnessing the inherent capabilities of transformer when compared with feature-based fusion approaches. Utilizing a feature-based fusion strategy, in which features from initial stages are processed by CNN or MLP networks and then merged with outputs of subsequent stages, results in an increase in the model's parameter count and a decline in performance, as demonstrated in the 2nd and 3rd rows of the table \ref{table:cross}. For the attention-based information fusion method, we explored various fusion strategies. Pyramidal XLR refers to initiating attention fusion simultaneously from the earliest and final stages towards the middle. Pyramidal $Q$ strategy introduces a linear layer in the final layer of the TT, modifying the output's temporal dimension and creating a pyramidal configuration for each module's $Q$. Residual XLR involves attention fusion between adjacent modules. The adjacent connection strategy emerged as the most effective through this experiment.

\begin{figure}[!t]
    \centering
    \centerline{\includegraphics[width=1.0\linewidth]{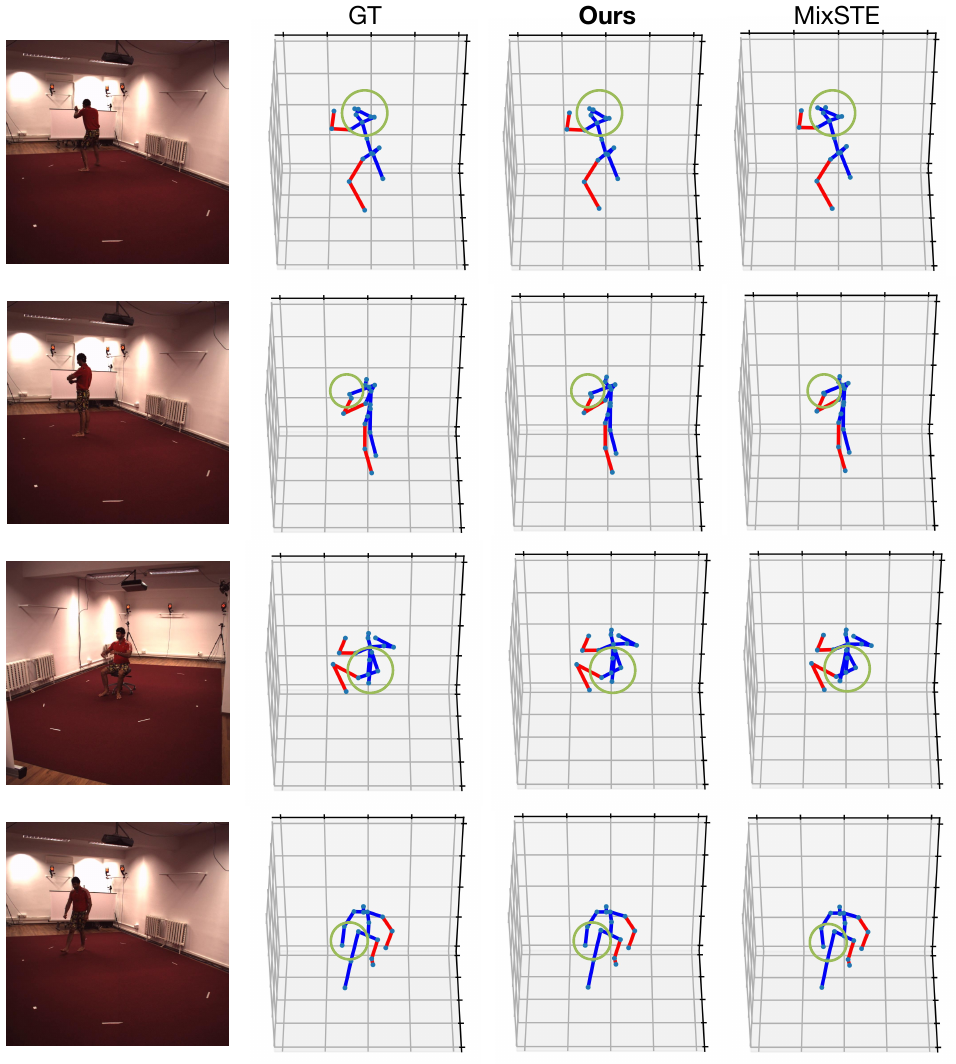}}
    \caption{\textbf{Comparison of 3D estimated human pose generated by different methods.} The 3D reconstruction visualization results using our proposed method, SOTA method MixSTE, ground truth, and the corresponding video frame in the Human3.6M dataset are shown in this figure. Our method shows higher accuracy and robustness in handling various actions and occlusion scenarios.}
    \label{fig:visualization}
\end{figure}

\subsection{Visualization and Qualitative Results}

Figure \ref{fig:visualization} provides a compelling comparison, highlighting the superiority of our RTPCA method over the current SOTA, MixSTE. This visualization clearly demonstrates our model's ability to generate highly accurate human poses, thereby confirming its effectiveness across a wide range of actions. Notably, our model exhibits remarkable performance in complex scenarios, particularly those involving occlusions, as depicted in the illustrative examples in the 1st and 2nd rows. These observations underscore the robustness and precision of our approach in capturing complex human poses.

\section{Conclusion}
In this paper, we introduce the RTPCA transformer, a novel framework for enhancing temporal modeling in 3D
Human Pose Estimation. Our framework employs a pyramidal compression-and-amplification structure to boost the intra-block temporal information fusion while incorporating a cross-layer attention mechanism to enhance inter-block attention interaction. Through the TPCA module, our framework captures multi-scale temporal features, enriching the representation of keys and values. These modules are connected via the XLR module, which facilitates dynamic interactions among queries, keys, and values. This interaction combines the early-stage information with the current stage to achieve superior accuracy and stability compared to other transformer-based baselines. 

\bibliographystyle{named}
\bibliography{ijcai24}

\clearpage
\section{Appendix}

\subsection{Temporal Stability Analysis}
\begin{figure}[h]
	\centering
	\centerline{\includegraphics[width=0.9\linewidth]{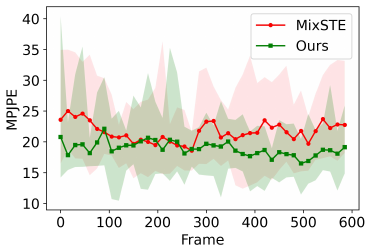}}
	\caption{ \textbf{The MPJPE-Frame curves using MixSTE and our methods. } The comparison of the proposed method and MixSTE on Human3.6M test set using frame-wise MPJPE for \textit{Photo} action is conducted and our method outperforms MixSTE with higher accuracy and stability.}
	\label{fig:error}
\end{figure}

The proposed \textbf{R}efined \textbf{T}emporal \textbf{P}yramidal \textbf{C}ompression-and-\textbf{A}mplification (RTPCA) transformer effectively mines the intra\&inter-block information from the attention mechanism, enhancing the model's robustness compared to methods solely employing stacked spatial-temporal transformers. The cross-layer refinement module incorporated into our model fosters a deeper interaction, acting as a highly innovative fusion strategy. To assess the enhancements made on the temporal transformer, we performed a temporal stationary analysis experiment. As depicted in Figure \ref{fig:error}, we measured the frame-wise Mean Per Joint Position Error (MPJPE) on the Human3.6M test set, with subjects S9 and S11 performing the \textit{Photo} action. The frame-wise MPJPE represents the average error of all estimated joints in each frame. The shaded area in the figure signifies the range of maximum and minimum values across the test samples, with the middle line illustrating the average. The green and red lines represent the MPJPE over 600 frames for our method and the current SOTA method (MixSTE), respectively. The results demonstrate that our method exhibits less fluctuation and lower MPJPE values, showcasing its stability and accuracy over the MixSTE method.

\subsection{Attention Visualization} 
\begin{figure}[h]
	\centering
	\centerline{\includegraphics[width=0.9\linewidth]{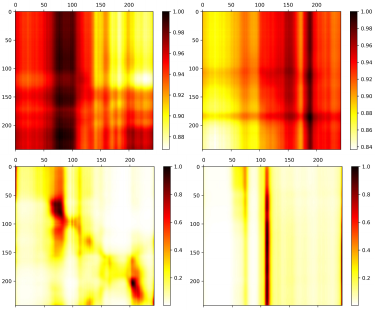}}
	\caption{ \textbf{Attention Visualization for our method and MixSTE.} The first row is the results of our method and the second row is the results of MixSTE. More comprehensive attention can be learned using our method which testifies the effectiveness to aggregate information. The original ST tends to focus on certain frames.}
	\label{fig:attn}
\end{figure}

To validate that the proposed RTPCA is capable of efficiently utilizing connections for information aggregation and exploring the attention mechanism, we conducted an attention visualization experiment, as demonstrated in Figure \ref{fig:attn}. Our model and MixSTE were evaluated on the Human3.6M test set S11 with the \textit{Sitting} action, and the attention matrix of the second temporal transformer was visualized. The first and second rows of Figure \ref{fig:attn} represent the visualization results of our method and MixSTE, respectively. Each attention matrix has a dimension of $F \times F$ ($F=243$), and this heatmap reflects the weighted relationship between every frame. Brighter colors signify higher attention scores, which are normalized in this figure. Through RTPCA, a more comprehensive understanding of attention can be obtained, whereas the original ST seems to focus predominantly on specific frames.

\end{document}